\documentclass[10pt,twocolumn,letterpaper]{article}

\usepackage{iccv}              

\definecolor{iccvblue}{rgb}{0.21,0.49,0.74}
\usepackage[pagebackref,breaklinks,colorlinks,allcolors=iccvblue]{hyperref}
\usepackage{graphicx}
\usepackage{enumitem}
\usepackage{amsmath}
\usepackage{tabularx}
\usepackage{booktabs}
\usepackage{multirow}
\usepackage[ruled,vlined]{algorithm2e}

\def\paperID{*****} 
\def\confName{ICCV}
\def\confYear{2025}

\title{GenEscape: Hierarchical Multi-Agent Generation of Escape Room Puzzles}
\author{
Mengyi Shan \quad
Brian Curless \quad
Ira Kemelmacher-Shlizerman \quad
Steve Seitz \\
University of Washington \\
Seattle, WA, USA \\
{\tt\small \{shanmy, curless, kemelmi, seitz\}@cs.washington.edu}
}

\begin{document}
\maketitle

\begin{abstract}
We challenge text-to-image models with generating escape room puzzle images that are visually appealing, logically solid, and intellectually stimulating. While base image models struggle with spatial relationships and affordance reasoning, we propose a hierarchical multi-agent framework that decomposes this task into structured stages: functional design, symbolic scene graph reasoning, layout synthesis, and local image editing. Specialized agents collaborate through iterative feedback to ensure the scene is visually coherent and functionally solvable. Experiments show that agent collaboration improves output quality in terms of solvability, shortcut avoidance, and affordance clarity, while maintaining visual quality. 

\end{abstract}

\section{Introduction}

Escape rooms are environments designed as interactive puzzles, where players must explore a confined scene, manipulate objects in a precise order, and ultimately exit the room. We challenge modern Vision-Language Models (VLMs) with the task of designing and building 2D escape room image puzzles. While those models produce aesthetically compelling images~\cite{openai2024gpt4o,deepmind2024gemini,wang2024cogvlm,li2024hunyuandit}, they struggle with complex scenes that require fine-grained spatial relationships, physical affordance reasoning, or multi-step functional coherence.

A well-designed escape room puzzle must satisfy two critical criteria: it must be solvable, meaning the affordances of objects form a coherent and logically sound sequence of actions; and it must provide sufficient visual cues that guide the player toward that intended solution. This requires not just accurate object placement but a deliberate visual design that supports human reasoning through spatial relationships and visual emphasis. Traditional vision-language models, lacking structured planning and feedback, often generate scenes that are visually plausible but fail to meet these standards—either missing key logical links or omitting visual signals needed to solve the puzzle.

\begin{figure}[t]
    \centering
    \includegraphics[width=\linewidth]{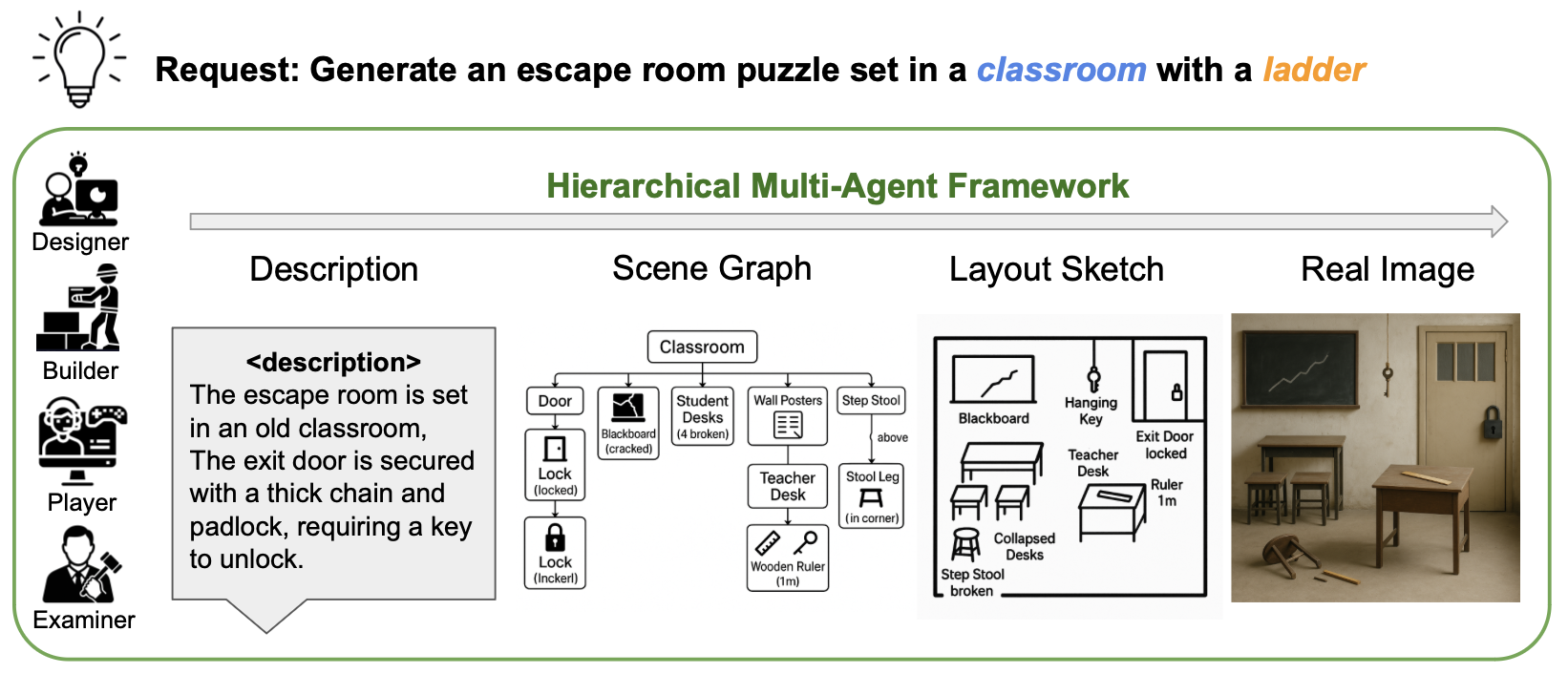}
    \caption{Four collaborative agents work together, hierarchically generating logically solid and visually appealing escape room puzzles through building scene graph, layout sketch towards photorealistic images. }
    \label{fig:teaser}
    
\end{figure}

\begin{figure*}
    \makebox[\textwidth][c]{%
        \includegraphics[width=\linewidth]{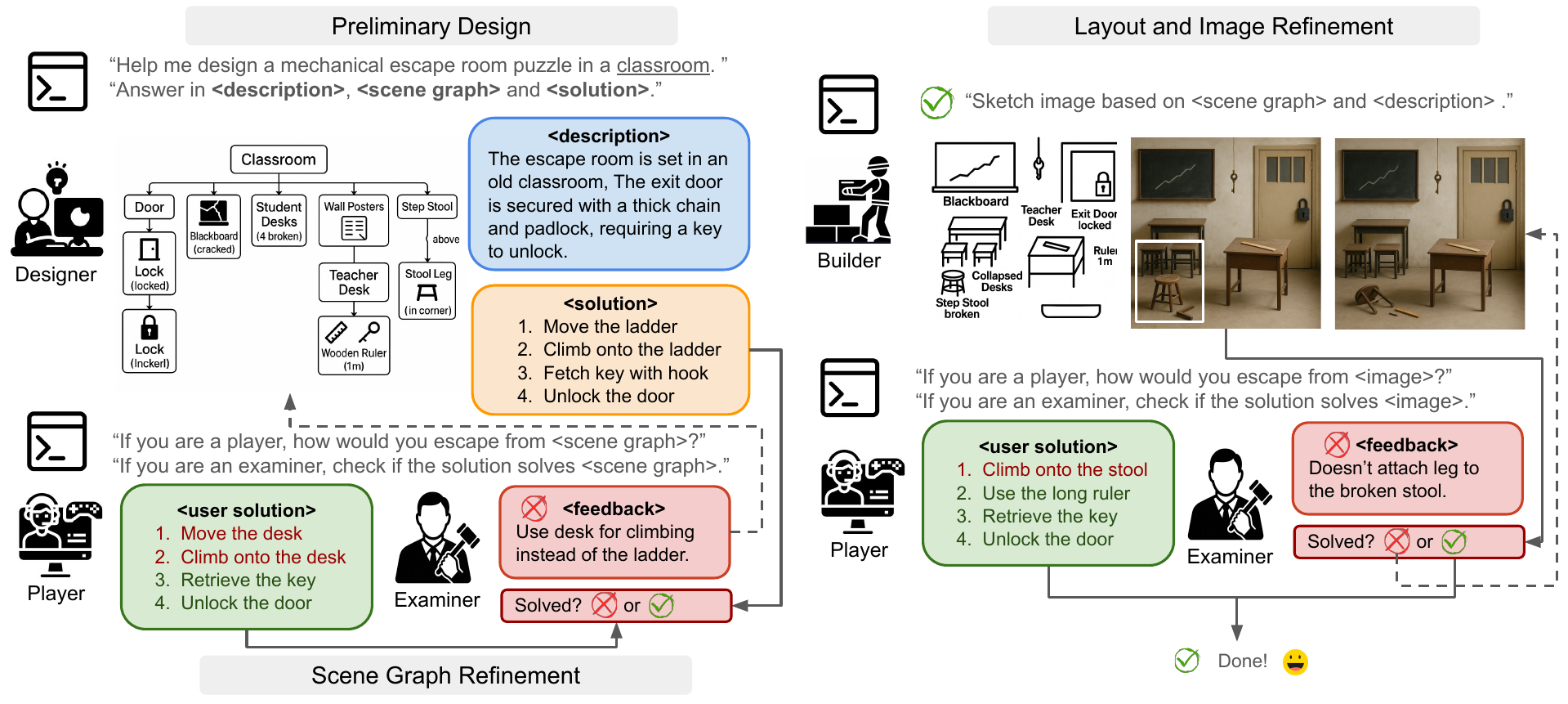}
    }
    \caption{Hierarchical multi-agent pipeline for escape room image generation. The \textit{Designer} creates an initial scene graph and solution. The \textit{Player} and \textit{Examiner} iteratively refine it for logical solvability. The \textit{Builder} then generates a 2D layout and image, which are further verified by the agents to ensure visual and functional consistency.}
    \label{fig:pipeline}
\end{figure*}

We propose a novel multi-agent ~\cite{zhou2024migc,neurips2024mllm,li2025mccd} interaction framework for escape room image generation. As in Fig~\ref{fig:teaser}, each agent contributes to a specific aspect of the scene (puzzle design, object placement, spatial consistency, or visual affordance verification) through iterative communication and refinement. This division of labor allows the system to reason about puzzle structure and object semantics in a modular way without sacrificing visual quality. 

Our contributions are summarized as follows:
\begin{itemize}[topsep=0pt,itemsep=2pt,parsep=0pt,partopsep=0pt]
    \item We propose the novel task of generating physical, photorealistic escape room puzzle images. 
    \item We propose a multi-agent collaborative system that hierarchically builds images through multiple levels of feedback on textual, symbolic and visual signals. 

\end{itemize}

\section{Related Work}
\noindent \textbf{Prompt Optimization.} Prompt optimization is a powerful strategy for guiding LLM without weight updates. In NLP, techniques such as \cite{shin2020autoprompt,lester2021power,ouyang2022training} show that well-crafted prompts improve downstream performance. Recent studies explore black-box prompt optimization~\cite{diao2023black,zhou2023evolutionary,park2023generative} for tasks like QA and reasoning. In vision, prompt engineering and multi-turn refinement help steer diffusion and vision-language models~\cite{liu2022compositional,hertz2022prompt,hao2023optimizing,mo2024dynamic,wu2024universal}. Our approach introduces a hierarchical framework with symbolic intermediates, reducing local minima risks and improving convergence.

\noindent \textbf{Multi-Agent Systems.}
Recent work has explored how multiple specialized agents can collaboratively solve complex tasks in language and vision~\cite{zhang2024reflective,xiong2025planning,google2024coa}. In visual generation, multi-agent setups have been applied to interactive storytelling, scene composition, and instruction-following environments, where agents assume distinct roles such as planning, verification, and rendering \cite{xu2024mmstoryagent, zhao2024lightva,hu2021scalable,gao2024llmabm}. 
Most prior work either focuses on interactive dialogue or relies on fixed procedural pipelines. 

\noindent \textbf{Generating Puzzles.} LLMs have been applied to puzzle generation and solving across multimodal domains. \citet{semeval2024ensemble} combined chain-of-thought prompting with LLMs for word and sentence puzzles. \citet{puzzlevqa2024} introduced a diagnostic benchmark of abstract visual puzzles to assess multimodal reasoning. \citet{neurips2024puzzles} proposed a reinforcement learning benchmark targeting logic-based algorithmic reasoning. EscapeCraft~\cite{wang2025multimodallargelanguagemodels} also focuses on escape room generation, but relies on 3D assets, differing from our 2D image-based approach.

\section{Method}
\noindent \textbf{Problem Formulation.}
We aim to generate an image of a mechanical escape room puzzle that is both visually realistic and logically solvable. Given a scene type keyword, a list of objects, and optionally a solution length $l$, the system produces an image $I$ and an intended solution $S$ (a list of action steps, each represented as a sentence of text) such that: (1) all key objects are present and spatially arranged to support a valid interaction sequence of at most $l$ steps, and (2) the visual cues embedded in the scene are sufficient to guide a player toward inferring the intended solution $S$ without shortcuts.

\subsection{Hierarchical Refinement Framework}
Our framework comprises four \textit{agents}, which are independent VLM instances assigned specific roles and communicating via text and visuals. The \textbf{Designer} generates the scene description, graph, and solution; the \textbf{Player} simulates a human solver; the \textbf{Examiner} compares the Player’s actions with the official solution and suggests refinements; and the \textbf{Builder} creates a 2D layout and photorealistic image aligned with the intended logic.

We adopt a hierarchical refinement strategy across four stages: text description, symbolic scene graph, 2D layout, and photorealistic image. At each stage, the \textit{Player} proposes a solution, and the \textit{Examiner} verifies solvability. This staged process reduces computational cost compared to direct image optimization while preserving object relationships and functional logic. We repeat until the \textit{Examiner} confirms the solution matches the official one. See Algorithm~\ref{alg:algorithm} for details.

\subsection{Preliminary Design}
In the first stage of our framework, the \textit{Designer} agent is prompted to generate three aligned outputs: (1) a scene description in natural language that sets up the environment and puzzle premise, (2) a structured scene graph tree in \texttt{yaml} format where each node is an object in the scene, and each pair of parent-child represents a spatial connection relationship, and (3) a solution sequence consisting of valid player actions that logically lead to unlocking the exit. Here, a “valid” action sequence is one that respects the physical constraints and object states specified in the scene graph—e.g., only using provided tools, manipulating reachable objects, and avoiding shortcuts or physically implausible steps.

\subsection{Scene Graph Optimization}
In this stage, we refine the scene graph purely through symbolic reasoning, without involving any visual modalities. The \textit{Player} agent attempts to solve the puzzle by generating a sequence of actions based solely on the scene graph structure. This proposed solution is then evaluated by the \textit{Examiner} agent, who compares it against the intended solution. The \textit{Examiner} highlights key discrepancies—such as when the player exploits an unintended shortcut—and summarizes them in bullet-point feedback. In response, the \textit{Examiner} revises the scene graph to eliminate such shortcuts and reinforce the intended solution path. Through iterative refinement, the scene graph is progressively updated until it supports a coherent, logically sound strategy that aligns with the ground-truth solution. This symbolic verification step ensures that downstream image generation faithfully captures a functionally valid and solvable puzzle.

\begin{algorithm}
\caption{Hierarchical Puzzle Optimization with Multi-Agent Feedback}
\SetKwInOut{Input}{Input}
\SetKwInOut{Output}{Output}
\Input{Initial scene graph $G_0$, ground-truth solution $S$}
\Output{Final image $I$ that supports $S$}

$R \leftarrow G_0$\;

\For{stage $t \in \{\textsc{Graph}, \textsc{Layout}, \textsc{Image}\}$}{
    \Repeat{$\Delta = \emptyset$}{
        $S^* \leftarrow \mathsf{player}.\mathsf{solve}(R)$\;
        $\Delta \leftarrow \mathsf{examiner}.\mathsf{check}(S, S^*)$\;
        \If{$t = \textsc{Graph}$}{
            $R \leftarrow \mathsf{examiner}.\mathsf{refine}(R, \Delta)$\;
        }
        \If{$t = \textsc{Layout}$}{
            $R \leftarrow \mathsf{builder}.\mathsf{refine}(R, \Delta)$\;
        }
        \If{$t = \textsc{Image}$}{
            $R \leftarrow \mathsf{builder}.\mathsf{refine}(R, \Delta, R_{\textsc{L}})$\;
        }
    }
}
\Return $R$\;

\label{alg:algorithm}

\end{algorithm}

\begin{figure*}[tp]
    \centering
    \includegraphics[width=\linewidth]{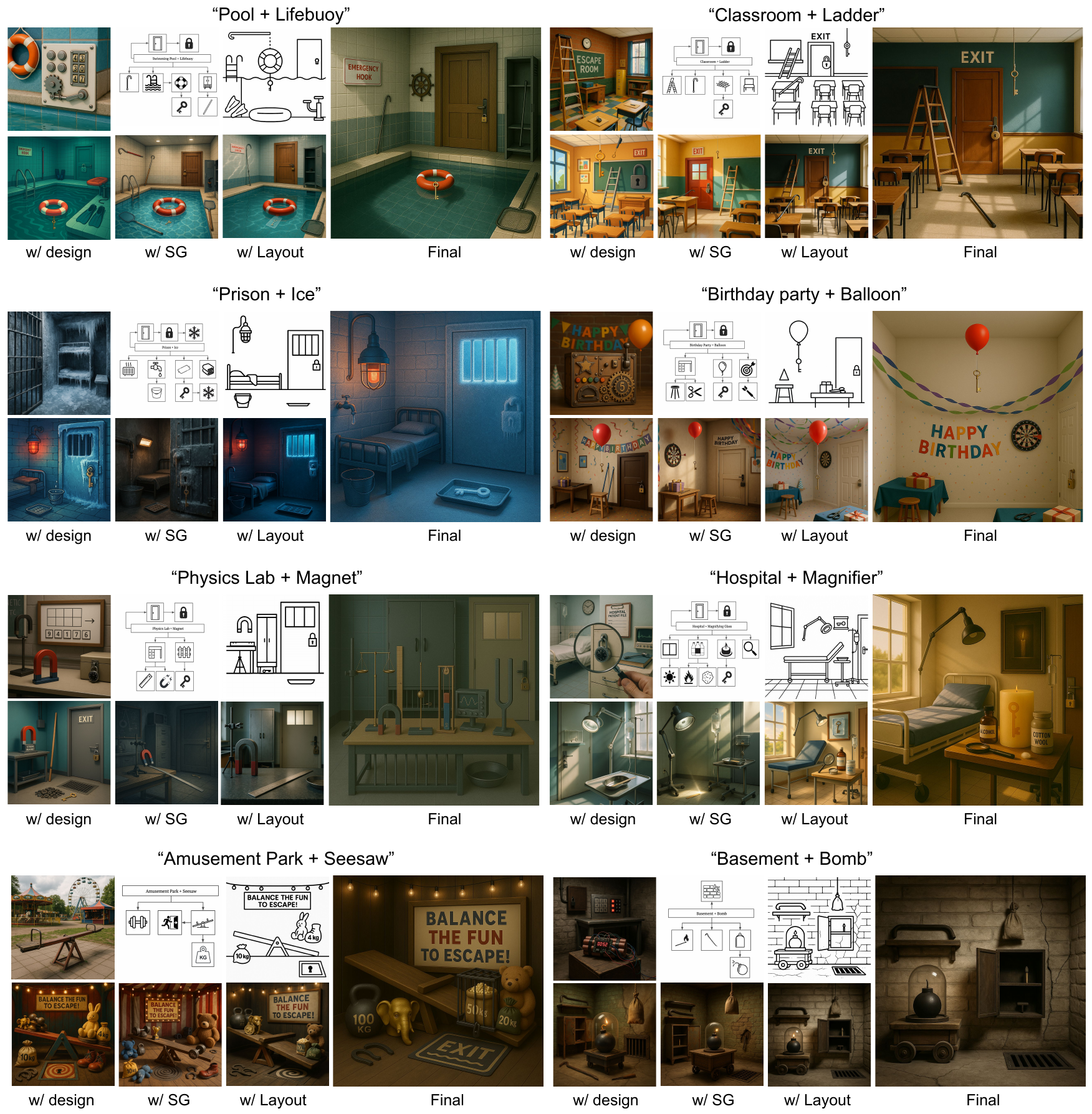}
    \caption{Our visual results compared with baseline and alternative designs. Left columns are naive generation (top) and assisted by a textual design (bottom). The second and third columns show the scene graph and 2D layout (top), and results with them (bottom). Right columns are the final results after visual signal-based image optimization. See if you can solve the puzzle! Official solutions and example human interaction are provided in the appendix.}
    \label{fig:result}
\end{figure*}

\subsection{Affordance-Guided Local Editing}
After the scene graph is finalized, we prompt the \textit{Builder} to generate a 2D layout (Fig. ~\ref{fig:pipeline} right) where each object is represented by an abstract icon positioned according to the spatial relationships in the scene graph. The same \textit{Player–Examiner} interaction loop is reused to guarantee that the layout visually supports the intended puzzle solution until convergence. 

Once the layout is verified, the \textit{Builder} renders a photorealistic image based on the layout and the description. In this final stage, the \textit{Examiner} again compares the \textit{Player}’s actions—now based on the image—to the intended solution. If the \textit{Player} misinterprets object affordances, the \textit{Examiner} identifies which visual cues are lacking in the image. Using the icon position from the layout stage, we then apply local image editing to enhance or suppress affordances to steer perception toward the correct interaction.

\section{Experiment}
We use GPT-4o ~\cite{openai2024gpt4o} API as our base model and generate square images at 1024 $\times$ 1024 resolution with the highest quality. This can be effectively transferred to any blackbox text-to-image model without any specific requirements on model architecture.

\subsection{Evaluation Metrics}
We evaluate our method using a combination of human judgments and automated metrics. Specifically, we consider three key human evaluation criteria. \textbf{Solvability} measures whether a player can infer the intended multi-step solution purely from visual cues in the generated image. \textbf{Shortcut Avoidance} assesses whether the image successfully prevents trivial or unintended solutions that bypass the intended logic. \textbf{Spatial Alignment} evaluates how accurately the visual scene reflects the object relations specified in the initial design, such as positions and reachability.

In addition to human assessments, we report the LongCLIP~\cite{zhang2024longclip} score, which quantifies the semantic alignment between the \textit{Designer} prompt and the final image, and the average number of image API calls used throughout the generation process (note that this metric is not applicable to baseline methods that lack image-level refinement). Evaluation is conducted on 15 diverse scene settings, each involving 2 core interactive objects. For each case, we collect responses from 10 human annotators, who are shown multiple generated images (from our method and baselines) along with the scene description and object names. They are asked to select the best image for each of the three criteria. We report the percentage of cases where each method's output is selected as the top choice.

\subsection{Analysis}We compare against vanilla GPT-4o~\cite{openai2024gpt4o} and several ablated versions of our framework by selectively removing refinement stages on the scene graph, layout sketch, and final image. Qualitative results are shown in Fig~\ref{fig:result}. For example, in the top-right classroom scene with the ladder, naively prompting GPT for an ``escape room'' (top left) yields a scene with generic visual cues but no coherent solution path. Providing only a textual description introduces some relevant elements but lacks spatial grounding (e.g., a lock floating on the blackboard). Incorporating the scene graph and layout planning significantly improves spatial coherence, aligning the visual arrangement with the intended interaction sequence. Final image refinement further removes visual artifacts (e.g., malformed hooks) and eliminates shortcuts (e.g., climbable desks) that could compromise puzzle integrity.

Quantitative results in Tab~\ref{tab:comparison_results} reinforce these observations. Starting from vanilla GPT-4o, which achieves only $3.3\%$ solvability and fails entirely on shortcut avoidance, each added module brings measurable gains. The scene graph (+D+S.G.'') more than quadruples shortcut avoidance to $13.3\%$ and introduces modest gains in spatial alignment ($26.7\%$). The addition of layout planning (+D+S.G.+L'') leads to further improvements in both solvability ($10.0\%$) and shortcut avoidance ($20.0\%$), while reducing the average number of required generations. Image-level refinement (``+D+S.G.+I'') helps further suppress unintended interactions and enhances alignment, though at the cost of slightly lower CLIP scores, likely due to localized editing that deviates from the original prompt.

Our full method, combining all stages, achieves the highest scores across all human evaluation metrics: $53.3\%$ for solvability, $46.6\%$ for shortcut avoidance, and $36.7\%$ for spatial alignment—substantially outperforming all baselines. Notably, it achieves these gains with far fewer image generations (4.5 on average), underscoring the efficiency benefits of our structured refinement process. Although CLIP scores slightly decline as editing introduces affordance cues not mentioned in the initial prompt, this trade-off still ensures functional solvability and visual clarity.

\begin{table}[t]
\centering
\small

\caption{Comparison of adding different stages to the vanilla GPT-4o model, including description (D), Scene Graph (S.G.), Layout (L), and Image editing (I). The last row is our full pipeline with all modules added.}
\begin{tabular}{p{1.3cm}p{0.8cm}p{0.8cm}p{0.8cm}p{0.8cm}p{0.9cm}}
\toprule
\textbf{Method} & \textbf{Solv.} & \textbf{Short.} & \textbf{Align.} & \textbf{CLIP} & \textbf{\#Gen.} \\
\midrule
GPT-4o & 3.3\% & 0.0\% & N/A & N/A & N/A \\
$+$D & 6.7\% & 3.3\% & 0.0\% & \textbf{0.42} & N/A \\
$+$D$+$S.G. & 6.7\% & 13.3\% & 26.7\% & 0.37 & N/A \\
$+$D$+$S.G.$+$L & 10.0\% & 20.0\% & 13.3\% & 0.38 & 13.2\\
$+$D$+$S.G.$+$I & 20.0\% & 16.7\% & 23.3\% & 0.32 & 15.8\\
\midrule
\textbf{Ours full} & \textbf{53.3\%} & \textbf{46.6\%} & \textbf{36.7\%} & 0.32 & \textbf{4.5} \\
\bottomrule
\end{tabular}
\label{tab:comparison_results}
\end{table}

\section{Human Interface}

\begin{figure}[t]
    \centering
    \includegraphics[width=0.9\linewidth]{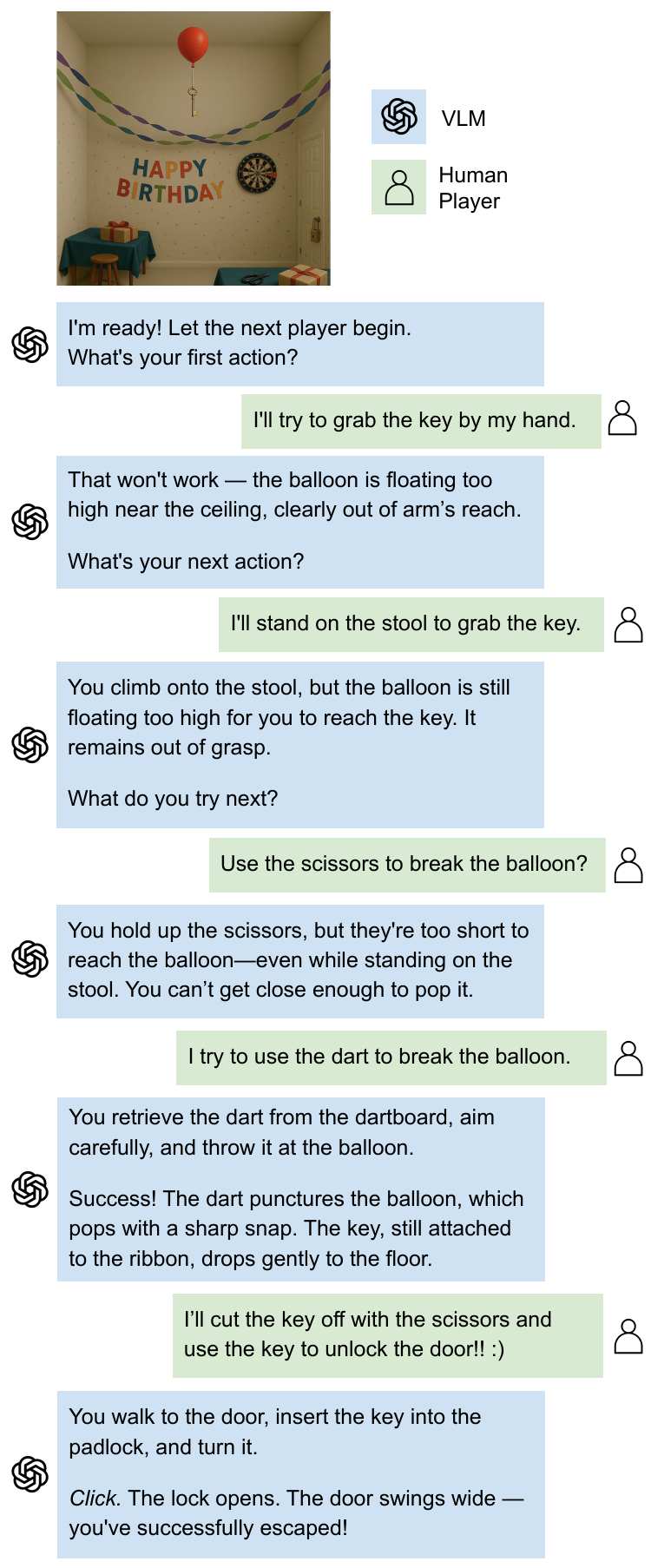}
    \caption{Real case of human-agent interaction. The human is typing in possible solutions trying to solve the escape room puzzle, while the AI is providing feedback and gradually guiding the player towards the ideal solution. We put them into specific figuring format for better spatial fitting. All texts are direct input and outputs of the VLM conversation.}
    \label{fig:interface-1}
\end{figure}

We develop an interactive human interface that allows users to engage with the generated escape room puzzles by attempting to solve them visually. Alongside the player, an AI agent—equipped with the ground-truth solution—evaluates the human's proposed actions, providing feedback on whether each step aligns with the intended logic. When discrepancies arise, the agent offers corrective guidance, helping the player iteratively converge toward the correct solution. This setup enables real-time puzzle solving and highlights the interpretability and playability of our generated scenes. See Fig.\ref{fig:interface-1} for a walkthrough of the interactive gameplay process.

\section{Conclusion}
We presented GenEscape, a hierarchical multi-agent framework for generating 2D escape room puzzles that are both visually realistic and logically solvable. By decomposing the generation process into symbolic reasoning, layout planning, and affordance-aware image refinement, our system ensures functional coherence and visual clarity at each stage. Specialized agents collaborate through iterative feedback loops to simulate human solving and verification, improving puzzle solvability and blocking unintended shortcuts. Experiments across diverse scenes show that our approach significantly outperforms vanilla generation and ablations in both human and automatic evaluations. This work highlights the value of structured, multi-agent coordination for complex visual reasoning tasks.

\section{Limitations}
While our hierarchical multi-agent framework effectively generates escape room puzzles that are solvable and visually coherent, it has several limitations. First, our current design supports only puzzles with fully visible objects. Players cannot interact with the environment to uncover hidden items (e.g., opening boxes or inspecting drawers), limiting the depth and realism of puzzle mechanics. Second, our model doesn't support very long solution chains, specifically, more than eight steps or eight objects involved. It makes mistakes on spatial layouts and takes a very long time to converge to a solvable image. Finally, we imagine that it would be cool if we can generate images of the resulting scene condition after each human step, but GPT-4o's limited ability on image editing still prevents us from creating perfectly aligned scene images before and after a human action.

\section*{Solution to Figure~\ref{fig:result}}

\subsection*{Pool + Lifebuoy}
\begin{enumerate}
    \item \textbf{Retrieve the skimmer net from the deck.} Pick up the lightweight, fully extended telescoping skimmer net leaning beside the pool.

    \item \textbf{Get the clamp from the locker.} Open the nearby locker and take the waterproof clamp with rubber-lined jaws.

    \item \textbf{Attach the clamp to the skimmer net.} Fasten the clamp securely to the end of the skimmer net, creating a long grabbing tool.

    \item \textbf{Use the extended tool to reach the lifebuoy.} Carefully extend the modified skimmer over the pool and use the clamp to grab the plastic pouch taped to the underside of the floating lifebuoy.

    \item \textbf{Unlock the exit door.} Open the sealed pouch, take out the brass key, and use it to unlock the padlocked door to escape the room.
\end{enumerate}

\subsection*{Classroom + Ladder}
\begin{enumerate}
    \item \textbf{Pick up the hooked pole from the teacher’s desk.} Retrieve the small pole with a hook from the surface of the desk at the front of the room.

    \item \textbf{Position the ladder beneath the key.} Move the ladder from its position near the chalkboard and place it safely under the dangling key.

    \item \textbf{Climb the ladder and use the hooked pole.} Ascend the ladder carefully and use the hooked pole to pull the key string within reach.

    \item \textbf{Grab the key and descend.} Once the key is close enough, grasp it securely and climb back down the ladder.

    \item \textbf{Unlock the exit door.} Insert the key into the padlock on the exit door and turn it to unlock and escape the room.
\end{enumerate}

\subsection*{Prison + Ice}
\begin{enumerate}
    \item \textbf{Place the metal bucket under the dripping faucet in the corner.} Allow it to collect a sufficient amount of water from the steady drip.

    \item \textbf{Move the filled bucket beneath the heat lamp.} Let it sit for several minutes so the water gradually warms up from the lamp's heat.

    \item \textbf{Tear off a large piece of the wool blanket and soak it in the warm water.} The thick fabric will retain both heat and moisture, making it ideal for melting ice.

    \item \textbf{Wrap the soaked section of blanket tightly around the frozen lock and key on the door.} Ensure it covers the area where the key is embedded in the ice to maximize thermal contact.

    \item \textbf{Wait for the heat from the blanket to melt the ice.} Once the ice has sufficiently thawed, pull out the key.

    \item \textbf{Use the key to unlock the door and escape the room.}
\end{enumerate}

\subsection*{Birthday Party + Balloon}
\begin{enumerate}
    \item \textbf{Pick up the scissors from the table.} Retrieve the visible scissors from the lower-right table.

    \item \textbf{Throw the dart at the balloon.} Take the dart from the dartboard and throw it at the red balloon to pop it or weigh it down.

    \item \textbf{Cut the string to retrieve the key.} Once the balloon is brought down and within reach, use the scissors to cut the string and release the brass key.

    \item \textbf{Use the key to unlock the door.} Insert the key into the lock on the door handle and turn it to escape the room.
\end{enumerate}

\subsection*{Physics Lab + Magnet}
\begin{enumerate}
    \item \textbf{Use the metal clamp to secure the magnet.} Attach the U-shaped magnet to the clamp stand so it can be held steadily and extended toward the cage.
    
    \item \textbf{Slide the magnet through the cage bars.} Position the magnet through the dense cage bars, aligning it with the brass key inside the cage.
    
    \item \textbf{Attract the key using the magnet.} Carefully maneuver the magnet to make contact with the metallic key and pull it toward the bars.

    \item \textbf{Retrieve the key from the cage.} Once the key is close enough to the bars, rotate and tilt the magnet to drag the key through a gap. Take the key out and use it to unlock the door.
\end{enumerate}

\subsection*{Hospital + Magnifier}

\begin{enumerate}
    \item \textbf{Soak the cotton wool with alcohol.} Pour a small amount of alcohol from the labeled bottle onto the cotton wool to make it flammable.
    
    \item \textbf{Place the soaked cotton wool on top of the candle.} This sets up an easy ignition point to begin melting the wax.
    
    \item \textbf{Use the magnifying glass to focus sunlight onto the cotton wool.} Position the cotton on the desk so that sunlight from the window can be concentrated using the magnifier. Adjust the angle until it ignites.
    
    \item \textbf{Let the flame melt the wax and retrieve the key.} As the wax burns and softens, the embedded key becomes accessible. Use it to unlock the door and exit the room.
\end{enumerate}
    
\subsection*{Amusement Park + Seesaw}
\begin{enumerate}
    \item \textbf{Read the sign for the puzzle hint.} Observe the message "BALANCE THE FUN TO ESCAPE!" to infer that equalizing the seesaw is the key to opening the exit.
    \item \textbf{Inspect the right side of the seesaw.} Notice that it contains immovable 50kg and 20kg weights locked inside a cage, indicating a fixed total weight on that side.
    
    \item \textbf{Collect available weights from the left side.} Find a 100kg kettlebell and a golden elephant head placed nearby on the ground.
    
    \item \textbf{Balance the seesaw using the collected items.} Carefully place the kettlebell and elephant head on the left side of the seesaw to counterbalance the fixed weights.
    
    \item \textbf{Trigger the hidden mechanism.} Once the seesaw achieves balance, pressure on the right end lifts and reveals a trapdoor labeled “EXIT”.
    
    \item \textbf{Escape through the trapdoor.} Pull the handle on the EXIT mat to open the trapdoor and leave the room.
    
\end{enumerate}

\subsection*{Basement + Bomb}
\begin{enumerate}
\item \textbf{Retrieve the crowbar.} Take the crowbar hanging from a wall-mounted shelf.
\item \textbf{Break the glass dome protecting the bomb.} Use the crowbar to smash the enclosure and gain access to the bomb.

\item \textbf{Find and collect the matchstick.} Look inside an open cabinet nearby and take the matchstick stored inside.

\item \textbf{Ignite the bomb's fuse.} Light the fuse using the matchstick to initiate the detonation sequence.

\item \textbf{Wait for the explosion to create an exit.} The bomb detonates and blows a hole through the weakened section of the basement wall.

\item \textbf{Escape through the destroyed wall.} Walk through the opening created by the explosion to exit the room.
\end{enumerate}

\newpage
{
    \small
    \bibliographystyle{ieeenat_fullname}
    \bibliography{main}

\begin{thebibliography}{30}
\providecommand{\natexlab}[1]{#1}
\providecommand{\url}[1]{\texttt{#1}}
\expandafter\ifx\csname urlstyle\endcsname\relax
  \providecommand{\doi}[1]{doi: #1}\else
  \providecommand{\doi}{doi: \begingroup \urlstyle{rm}\Url}\fi

\bibitem[Chia et~al.(2024)Chia, Toh, Ghosal, Bing, and Poria]{puzzlevqa2024}
Yew~Ken Chia, Vernon Toh, Deepanway Ghosal, Lidong Bing, and Soujanya Poria.
\newblock Puzzlevqa: Diagnosing multimodal reasoning challenges of language
  models with abstract visual patterns.
\newblock In \emph{Findings of the Association for Computational Linguistics:
  ACL 2024}, 2024.

\bibitem[DeepMind(2024)]{deepmind2024gemini}
Google DeepMind.
\newblock Gemini 1.5 technical report, 2024.
\newblock Accessed: 2025-05-08.

\bibitem[Diao et~al.(2023)Diao, Xu, Zhang, and Li]{diao2023black}
Qi Diao, Zhengxiao Xu, Tong Zhang, and Hang Li.
\newblock Black-box prompt optimization with meta reinforcement learning.
\newblock In \emph{International Conference on Learning Representations
  (ICLR)}, 2023.
\newblock arXiv:2201.08531.

\bibitem[Estermann et~al.(2024)Estermann, Lanzendörfer, Niedermayr, and
  Wattenhofer]{neurips2024puzzles}
Benjamin Estermann, Luca Lanzendörfer, Yannick Niedermayr, and Roger
  Wattenhofer.
\newblock Puzzles: A benchmark for neural algorithmic reasoning.
\newblock In \emph{Advances in Neural Information Processing Systems
  (NeurIPS)}, 2024.

\bibitem[Gao et~al.(2024)Gao, Lan, Li, Yuan, Ding, Zhou, Xu, and
  Li]{gao2024llmabm}
Chen Gao, Xiaochong Lan, Nian Li, Yuan Yuan, Jingtao Ding, Zhilun Zhou, Fengli
  Xu, and Yong Li.
\newblock Large language models empowered agent-based modeling and simulation:
  A survey and perspectives.
\newblock \emph{Humanities and Social Sciences Communications}, 11\penalty0
  (1):\penalty0 1259, 2024.

\bibitem[{Google Research}(2024)]{google2024coa}
{Google Research}.
\newblock Chain-of-agents: Large language models collaborating on long context
  tasks.
\newblock
  \url{https://research.google/blog/chain-of-agents-large-language-models-collaborating-on-long-context-tasks/},
  2024.

\bibitem[Hao et~al.(2023)Hao, Chi, Dong, and Wei]{hao2023optimizing}
Yaru Hao, Zewen Chi, Li Dong, and Furu Wei.
\newblock Optimizing prompts for text-to-image generation.
\newblock In \emph{Advances in Neural Information Processing Systems}, 2023.

\bibitem[Hertz et~al.(2023)Hertz, Mokady, Gal, Bermano, and
  Cohen-Or]{hertz2022prompt}
Amir Hertz, Ron Mokady, Rinon Gal, Amit~H Bermano, and Daniel Cohen-Or.
\newblock Prompt-to-prompt image editing with cross-attention control.
\newblock In \emph{International Conference on Learning Representations
  (ICLR)}, 2023.

\bibitem[Hu et~al.(2021)Hu, Gama, Chen, Zheng, Wang, Ribeiro, and
  Sadler]{hu2021scalable}
Ting-Kuei Hu, Fernando Gama, Tianlong Chen, Wenqing Zheng, Zhangyang Wang,
  Alejandro Ribeiro, and Brian~M Sadler.
\newblock Scalable perception-action-communication loops with convolutional and
  graph neural networks.
\newblock \emph{arXiv preprint arXiv:2106.13358}, 2021.

\bibitem[Lester et~al.(2021)Lester, Al-Rfou, and Constant]{lester2021power}
Brian Lester, Rami Al-Rfou, and Noah Constant.
\newblock The power of scale for parameter-efficient prompt tuning.
\newblock In \emph{Proceedings of the 2021 Conference on Empirical Methods in
  Natural Language Processing (EMNLP)}, pages 3045--3059, 2021.

\bibitem[Li et~al.(2025)Li, Hou, Liu, Yang, Qian, Chen, Wei, Jiang, Xu, and
  Zhang]{li2025mccd}
Mingcheng Li, Xiaolu Hou, Ziyang Liu, Dingkang Yang, Ziyun Qian, Jiawei Chen,
  Jinjie Wei, Yue Jiang, Qingyao Xu, and Lihua Zhang.
\newblock Mccd: Multi-agent collaboration-based compositional diffusion for
  complex text-to-image generation.
\newblock In \emph{Proceedings of the IEEE/CVF Conference on Computer Vision
  and Pattern Recognition (CVPR)}, 2025.

\bibitem[Li et~al.(2024)Li, Zhang, Lin, Xiong, Long, Deng, Zhang, Liu, Huang,
  Xiao, Chen, He, Li, Li, Zhang, Quan, Lu, Huang, Yuan, Zheng, Li, Zhang,
  Zhang, Chen, Liu, Fang, Wang, Xue, Tao, Zhu, Liu, Lin, Sun, Li, Wang, Chen,
  Hu, Xiao, Chen, Liu, Liu, Wang, Yang, Jiang, and Lu]{li2024hunyuandit}
Zhimin Li, Jianwei Zhang, Qin Lin, Jiangfeng Xiong, Yanxin Long, Xinchi Deng,
  Yingfang Zhang, Xingchao Liu, Minbin Huang, Zedong Xiao, Dayou Chen, Jiajun
  He, Jiahao Li, Wenyue Li, Chen Zhang, Rongwei Quan, Jianxiang Lu, Jiabin
  Huang, Xiaoyan Yuan, Xiaoxiao Zheng, Yixuan Li, Jihong Zhang, Chao Zhang,
  Meng Chen, Jie Liu, Zheng Fang, Weiyan Wang, Jinbao Xue, Yangyu Tao, Jianchen
  Zhu, Kai Liu, Sihuan Lin, Yifu Sun, Yun Li, Dongdong Wang, Mingtao Chen,
  Zhichao Hu, Xiao Xiao, Yan Chen, Yuhong Liu, Wei Liu, Di Wang, Yong Yang, Jie
  Jiang, and Qinglin Lu.
\newblock Hunyuan-dit: A powerful multi-resolution diffusion transformer with
  fine-grained chinese understanding, 2024.

\bibitem[Liu et~al.(2022)Liu, Li, Du, Torralba, and
  Tenenbaum]{liu2022compositional}
Shuang Liu, Song Li, Yilun Du, Antonio Torralba, and Joshua~B Tenenbaum.
\newblock Compositional visual generation with composable diffusion models.
\newblock In \emph{European Conference on Computer Vision (ECCV)}, 2022.

\bibitem[Mo et~al.(2024)Mo, Zhang, Bai, Su, Wen, and Yang]{mo2024dynamic}
Wenyi Mo, Tianyu Zhang, Yalong Bai, Bing Su, Ji-Rong Wen, and Qing Yang.
\newblock Dynamic prompt optimizing for text-to-image generation.
\newblock In \emph{Proceedings of the IEEE/CVF Conference on Computer Vision
  and Pattern Recognition}, 2024.

\bibitem[OpenAI(2024)]{openai2024gpt4o}
OpenAI.
\newblock Gpt-4o technical report, 2024.
\newblock Accessed: 2025-05-08.

\bibitem[Ouyang et~al.(2022)Ouyang, Wu, Jiang, and et~al.]{ouyang2022training}
Long Ouyang, Jeff Wu, Xu Jiang, and et al.
\newblock Training language models to follow instructions with human feedback.
\newblock In \emph{Advances in Neural Information Processing Systems
  (NeurIPS)}, 2022.

\bibitem[Park et~al.(2023)Park, O'Brien, Cai, Morris, Liang, and
  Bernstein]{park2023generative}
Joon~Sung Park, Joseph~C O'Brien, Carrie~J Cai, Meredith~Ringel Morris, Percy
  Liang, and Michael~S Bernstein.
\newblock Generative agents: Interactive simulacra of human behavior.
\newblock \emph{arXiv preprint arXiv:2304.03442}, 2023.

\bibitem[Raihan et~al.(2024)Raihan, Goswami, Bin~Emran, Puspo, Ganguly, and
  Zampieri]{semeval2024ensemble}
Md~Nishat Raihan, Dhiman Goswami, Al~Nahian Bin~Emran, Sadiya Sayara~Chowdhury
  Puspo, Amrita Ganguly, and Marcos Zampieri.
\newblock Solving puzzles with an ensemble of chain-of-thought prompts.
\newblock In \emph{Proceedings of the 18th International Workshop on Semantic
  Evaluation (SemEval)}, 2024.

\bibitem[Shin et~al.(2020)Shin, Razeghi, Logan~IV, Wallace, and
  Singh]{shin2020autoprompt}
Taylor Shin, Yasaman Razeghi, Robert~L Logan~IV, Eric Wallace, and Suchin
  Singh.
\newblock Autoprompt: Eliciting knowledge from language models with
  automatically generated prompts.
\newblock In \emph{Proceedings of the 2020 Conference on Empirical Methods in
  Natural Language Processing (EMNLP)}, pages 4222--4235, 2020.

\bibitem[Wang et~al.(2024{\natexlab{a}})Wang, Zhang, Tang,
  et~al.]{wang2024cogvlm}
Can Wang, Zhikang Zhang, Zhanming Tang, et~al.
\newblock Cogvlm: Visual expert for pretrained language models.
\newblock \emph{arXiv preprint arXiv:2403.07584}, 2024{\natexlab{a}}.

\bibitem[Wang et~al.(2024{\natexlab{b}})Wang, Li, Li, and Liu]{neurips2024mllm}
Zhenyu Wang, Aoxue Li, Zhenguo Li, and Xihui Liu.
\newblock Multimodal llm as an agent for unified image generation and editing.
\newblock In \emph{Advances in Neural Information Processing Systems
  (NeurIPS)}, 2024{\natexlab{b}}.

\bibitem[Wang et~al.(2025)Wang, Dong, Luo, Ruan, Cheng, Chen, Li, and
  Liu]{wang2025multimodallargelanguagemodels}
Ziyue Wang, Yurui Dong, Fuwen Luo, Minyuan Ruan, Zhili Cheng, Chi Chen, Peng
  Li, and Yang Liu.
\newblock How do multimodal large language models handle complex multimodal
  reasoning? placing them in an extensible escape game, 2025.

\bibitem[Wu et~al.(2024)Wu, Gao, Wang, Zhang, and Wang]{wu2024universal}
Zongyu Wu, Hongcheng Gao, Yueze Wang, Xiang Zhang, and Suhang Wang.
\newblock Universal prompt optimizer for safe text-to-image generation.
\newblock In \emph{Proceedings of the 2024 Conference of the North American
  Chapter of the Association for Computational Linguistics: Human Language
  Technologies}, pages 6340--6354, 2024.

\bibitem[Xiong et~al.(2025)Xiong, Cheng, Xu, et~al.]{xiong2025planning}
Yujie Xiong, Yuan Cheng, Yinghui Xu, et~al.
\newblock Planning with multi-constraints via collaborative language agents.
\newblock In \emph{Proceedings of the 2025 Conference on Computational
  Linguistics (COLING)}, 2025.

\bibitem[Xu et~al.(2024)Xu, Mei, Li, Wu, Yan, Lai, Zhang, and
  Wu]{xu2024mmstoryagent}
Xuenan Xu, Jiahao Mei, Chenliang Li, Yuning Wu, Ming Yan, Shaopeng Lai, Ji
  Zhang, and Mengyue Wu.
\newblock Mm-storyagent: Immersive narrated storybook video generation with a
  multi-agent paradigm across text, image and audio.
\newblock \emph{arXiv preprint arXiv:2503.05242}, 2024.

\bibitem[Zhang et~al.(2024{\natexlab{a}})Zhang, Zhang, Dong, Zang, and
  Wang]{zhang2024longclip}
Beichen Zhang, Pan Zhang, Xiaoyi Dong, Yuhang Zang, and Jiaqi Wang.
\newblock Long-clip: Unlocking the long-text capability of clip.
\newblock \emph{arXiv preprint arXiv:2403.15378}, 2024{\natexlab{a}}.

\bibitem[Zhang et~al.(2024{\natexlab{b}})Zhang, Bai, Jia,
  et~al.]{zhang2024reflective}
Zeyu Zhang, Xuefeng Bai, Zhiwei Jia, et~al.
\newblock Reflective multi-agent collaboration based on large language models.
\newblock In \emph{Advances in Neural Information Processing Systems
  (NeurIPS)}, 2024{\natexlab{b}}.

\bibitem[Zhao et~al.(2024)Zhao, Wang, Xiang, Zhang, Guo, Turkay, Zhang, and
  Chen]{zhao2024lightva}
Yuheng Zhao, Junjie Wang, Linbin Xiang, Xiaowen Zhang, Zifei Guo, Cagatay
  Turkay, Yu Zhang, and Siming Chen.
\newblock Lightva: Lightweight visual analytics with llm agent-based task
  planning and execution.
\newblock \emph{arXiv preprint arXiv:2411.05651}, 2024.

\bibitem[Zhou et~al.(2024)Zhou, Li, Ma, Zhang, and Yang]{zhou2024migc}
Dewei Zhou, You Li, Fan Ma, Xiaoting Zhang, and Yi Yang.
\newblock Migc: Multi-instance generation controller for text-to-image
  synthesis.
\newblock In \emph{Proceedings of the IEEE/CVF Conference on Computer Vision
  and Pattern Recognition (CVPR)}, 2024.

\bibitem[Zhou and Neubig(2023)]{zhou2023evolutionary}
Zihan Zhou and Graham Neubig.
\newblock Large language model guided evolutionary optimization for black-box
  prompt tuning.
\newblock \emph{arXiv preprint arXiv:2305.14216}, 2023.

\end{thebibliography}
}

\end{document}